%% file: main.tex
\documentclass{article} 
\usepackage{neurips_2020,times}

\usepackage{hyperref}
\usepackage{url}
\input{math_commands.tex}

\usepackage[utf8]{inputenc} 
\usepackage{booktabs}       
\usepackage{amsfonts}       
\usepackage{nicefrac}       
\usepackage{microtype}      
\usepackage{optidef}
\usepackage{xcolor}
\usepackage{dsfont}
\usepackage{amssymb}
\usepackage{paralist}
\usepackage{enumitem}
\input{inputs}
\definecolor{myblue}{RGB}{0, 51, 153}
\usepackage{algorithm}
\usepackage{algorithmic}

\usepackage{newfloat}
\usepackage{listings}
\usepackage{graphicx} 
\usepackage{natbib}
\usepackage{wrapfig}
\usepackage{caption}
\usepackage{mdframed}

\usepackage{bibentry}

\newcommand{\pbox}[3]{%
  \begingroup
  \setlength{\fboxsep}{#1} 
  \colorbox{#2}{$#3$}
  \endgroup
}

\title{Predict-Then-Optimize by Proxy: Learning Joint Models of Prediction and Optimization}

\author{
James Kotary \\
University of Virginia \\
jk4pn@virginia.edu\And 
Vincenzo Di Vito \\
University of Virginia \\
eda8pc@virginia.edu \And 
Jacob Christopher \\
University of Virginia \\
csk4sr@virginia.edu \And
Pascal Van Hentenryck \\
Georgia Institute of Technology \\
pvh@isye.gatech.edu \And 
Ferdinando Fioretto \\
University of Virginia \\
fioretto@virginia.edu
}

\begin{document}

\maketitle

\begin{abstract}
Many real-world decision processes are modeled by optimization problems whose defining parameters are unknown and must be inferred from observable data. 
The Predict-Then-Optimize framework uses machine learning models to predict unknown parameters of an optimization problem from features before solving. Recent works show that decision quality can be improved in this setting by solving and differentiating the optimization problem in the training loop, enabling end-to-end training with loss functions defined directly on the resulting decisions. However, this approach can be inefficient and requires handcrafted, problem-specific rules for backpropagation through the optimization step. This paper proposes an alternative method, in which optimal solutions are learned directly from the observable features by predictive models. The approach is generic, and based on an adaptation of the Learning-to-Optimize paradigm, from which a rich variety of existing techniques can be employed. Experimental evaluations show the ability of several Learning-to-Optimize methods to provide efficient, accurate, and flexible solutions to an array of challenging Predict-Then-Optimize problems.
\end{abstract}

\section{Introduction}
\label{sec:introduction}
The \emph{Predict-Then-Optimize} (PtO) framework models decision-making processes as optimization problems whose parameters are only partially known while the remaining, unknown, parameters must be estimated by a machine learning (ML) model. 
The predicted parameters complete the specification of an optimization problem which is then solved to produce a final decision. 
The problem is posed as estimating the solution $\bm{x}^\star(\zeta) \in \cX \subseteq \mathbb{R}^n$ of a \emph{parametric} optimization problem:
\begin{align}
    \label{eq:opt_generic}
    \bm{x}^\star(\bm{\zeta}) = \argmin_{\bm{x}} &\;\; f(\bm{x}, \bm{\zeta})    
    \\
    \textsl{such that:} &\;\;    \bm{g}(\bm{x}) \leq 0, \;\;   \bm{h}(\bm{x}) = 0, \notag
\end{align}
given that parameters $\bm \zeta \in \cC \subseteq \RR^p$ are unknown, but that a correlated set of observable values $\bm{z} \in \cZ$ are available. Here $f$ is an objective function, and $\bm g$ and $\bm h$ define the set of the problem's inequality and equality constraints. 
The combined prediction and optimization model is evaluated on the basis of the optimality of its downstream decisions, with respect to $f$ under its ground-truth problem parameters \citep{elmachtoub2020smart}. 
This setting is ubiquitous to many real-world applications confronting the task of decision-making under uncertainty, such as planning the shortest route in a city, determining optimal power generation schedules, or managing investment portfolios.
For example, a vehicle routing system may aim to minimize a rider's total commute time by solving a shortest-path optimization model (\ref{eq:opt_generic}) given knowledge of the transit times $\bm{\zeta}$ over each individual city block. In absence of that knowledge, it may be estimated by models trained to predict local transit times  based on exogenous data $\bm{z}$, such as weather and traffic conditions. In this context, more accurately predicted transit times $\hat{\bm{\zeta}}$ tend to produce routing plans $\bm{x}^{\star}({\hat{\bm{\zeta}}})$ with shorter overall commutes, with respect to the true city-block transit times $\bm{\zeta}$.

However, direct training of predictions from observable features to problem parameters tends to generalize poorly with respect to the ground-truth optimality achieved by a subsequent decision model \citep{mandi2023decision,kotary2021end}. 
To address this challenge, \emph{End-to-end Predict-Then-Optimize} (EPO) \citep{elmachtoub2020smart} has emerged as a transformative paradigm in data-driven decision making in which predictive models are trained by directly minimizing loss functions defined on the downstream optimal solutions $\bm{x}^{\star}({\hat{\bm{\zeta}}})$. 

On the other hand, EPO implementations require backpropagation through the solution of the optimization problem (\ref{eq:opt_generic}) as a function of its parameters for end-to-end training. The required back-\\pro\-pagation rules are highly dependent on\\[-10pt]
\begin{wrapfigure}[18]{r}{0.55\linewidth}
    \vspace{-26pt}
    \centering
    \includegraphics[width=\linewidth]{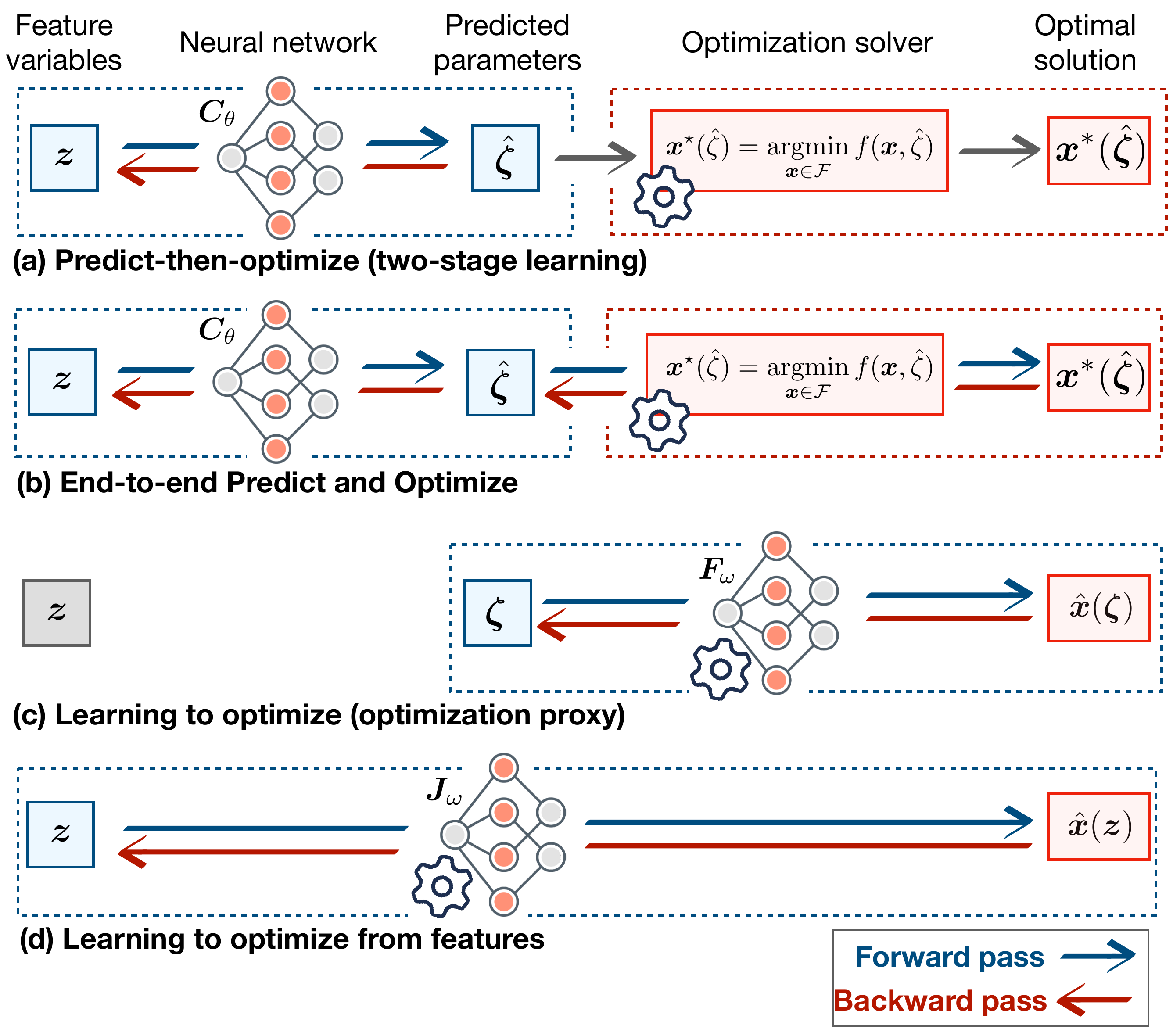}
    \vspace{-20pt}
    \caption{\small Illustration of Learning to Optimize from Features, in relation to other learning paradigms.}
    \label{fig:LtOF_diagram}
\end{wrapfigure}
the form of the optimization model and are typically derived by hand analytically for limited classes of models \citep{amos2019optnet,agrawal2019differentiable}. Furthermore, difficult decision models involving nonconvex or discrete optimization may not admit well-defined backpropagation rules. 

To address these challenges, this paper outlines a framework for training Predict-Then-Optimize models by techniques adapted from a separate but related area of work that combines constrained optimization end-to-end with machine learning. 
Such paradigm, called \emph{Learn-to-Optimize} (LtO), learns a mapping between the parameters of an optimization problem and its corresponding optimal solutions using a deep neural network (DNN), as illustrated in Figure \ref{fig:LtOF_diagram}(c). The resulting DNN mapping is then treated as an \emph{optimization proxy} whose role is to repeatedly solve difficult, but related optimization problems in real time \citep{vesselinova2020learning,fioretto2020lagrangian}. Several LtO methods specialize in training proxies to solve difficult problem forms, especially those involving nonconvex optimization. 

The proposed methodology of this paper, called \emph{Learning to Optimize from Features} (LtOF), recognizes that existing Learn-to-Optimize methods can provide an array of implementations for producing learned optimization proxies, which can handle hard optimization problem forms, have fast execution speeds, and are differentiable by construction. As such, they can be adapted to the Predict-Then-Optimize setting, offering an alternative to hard optimization solvers with handcrafted backpropagation rules. However, directly transferring a pretrained optimization proxy into the training loop of an EPO model leads to poor accuracy, as shown in Section \ref{sec:EPOwithProxy}, due to the inability of LtO proxies to generalize outside their training distribution. To circumvent this distributional shift issue, this paper shows how to adapt the LtO methodology to learn optimal solutions directly from features.

\textbf{Contributions.} In summary, this paper makes the following novel contributions: \textbf{(1)} It investigates the use of pretrained LtO proxy models as a means to approximate the decision-making component of the PtO pipeline, and demonstrates a distributional shift effect between prediction and optimization models that leads to loss of accuracy in end-to-end training. \textbf{(2)} It proposes Learning to Optimize from Features (LtOF), in which existing LtO methods are adapted to learn solutions to optimization problems directly from observable features, circumventing the distribution shift effect over the problem parameters. \textbf{(3)} The generic LtOF framework is evaluated by adapting several well-known LtO methods to solve Predict-then-Optimize problems with difficult optimization components, under complex feature-to-parameter mappings. Besides the performance improvement over two-stage approaches, \emph{the results show that difficult nonconvex optimization components can be incorporated into PtO pipelines naturally}, extending the flexibility and expressivity of PtO models.

\section{Problem Setting and Background}
\label{sec:PtO}

In the Predict-then-Optimize (PtO) setting, a (DNN) prediction model $\bm{C}_\theta: \cZ \to \cC \subseteq \RR^k$ first takes as input a feature vector $\bm{z} \in \cZ$ to produce predictions $\hat{\bm{\zeta}} = \bm{C}_\theta(\bm{z})$.  The model $\bm{C}$ is itself parametrized by learnable weights $\theta$. 
The predictions $\hat{\bm{\zeta}}$ are used to parametrize an optimization model of the form (\ref{eq:opt_generic}), which is then solved to produce optimal decisions $\bm{x}^\star(\hat{\bm{\zeta}}) \in \cX$. 
We call these two components, respectively, the \emph{first} and \emph{second} stage models. Combined, their goal is to produce decisions $\bm{x}^\star(\hat{\bm{\zeta}})$ which minimize the ground-truth objective value $f(\bm{x}^\star(\hat{\bm{\zeta}}), \bm{\zeta})$ given an observation of $\bm{z} \in \cZ$. Concretely, assuming a dataset of samples $(\bm{z}, \bm{\zeta})$ drawn from a joint distribution $\Omega$,  the goal is to learn a model $\bm{C}_\theta: \cZ \to \cC$ producing predictions $\hat{\bm{\zeta}} = \bm{C}_\theta (\bm{z})$ which achieves
\begin{equation}\label{eq:epo_goal}
    \minimize_\theta\; \EE_{(\bm{z}, \bm{\zeta}) \sim \Omega}\left[
        f\left( \bm{x}^\star(\hat{\bm{\zeta}}), \bm{\zeta} \right) 
        \right].
\end{equation}
This optimization is equivalent to minimizing expected \emph{regret}, defined as the magnitude of suboptimality of $\bm{x}^{\star}(\hat{\zeta})$ with respect to the ground-truth parameters: 
\begin{equation}\label{eq:regret_def}
\textit{regret}(\bm{x}^{\star}(\hat{\zeta}), \zeta) = f(\bm{x}^\star(\hat{\bm{\zeta}}), \bm{\zeta}) - f(\bm{x}^\star(\bm{\zeta}), \bm{\zeta}).
\end{equation}

\textbf{Two-stage Method.} A common approach to training the prediction model $\hat{\bm{\zeta}} = \bm{C}_\theta(\bm{z})$ is the \emph{two-stage} method, which trains to minimize the mean squared error loss 
$\ell(\hat{\bm{\zeta}}, \bm{\zeta}) = \| \hat{\bm{\zeta}} - \bm{\zeta} \|_2^2$, without taking into account the second stage optimization. While directly minimizing the prediction errors is confluent with the task of optimizing ground-truth objective 
$f(\bm{x}^\star(\hat{\bm{\zeta}}), \bm{\zeta})$, the separation of the two stages in training leads to  error propagation with respect to the optimality of downstream decisions, due to misalignment of the training loss with the true objective \citep{elmachtoub2020smart}. 

\textbf{End-to-End Predict-Then-Optimize.}
Improving on the two-stage method, the End-to-end Predict-end-Optimize (EPO) approach trains directly to optimize the objective  $f(\bm{x}^\star(\hat{\bm{\zeta}}), \bm{\zeta})$ by gradient descent, which is enabled by finding or approximating the derivatives through $\bm{x}^\star(\hat{\bm{\zeta}})$. This allows for end-to-end training of the PtO goal (\ref{eq:epo_goal}) directly as a loss function, 
which consistently outperforms two-stage methods with respect to the evaluation metric (\ref{eq:epo_goal}), especially when the mapping $\bm{z} \to \bm{\zeta}$ is \emph{difficult to learn} and subject to significant prediction error.
Such an integrated training of prediction and optimization is referred to as \emph{Smart Predict-Then-Optimize} \citep{elmachtoub2020smart}, \emph{Decision-Focused Learning} \citep{wilder2018melding}, or {End-to-End Predict-Then-Optimize} (EPO) \citep{tang2022pyepo}. This paper adopts the latter term throughout, for consistency. Various implementations of this idea have shown significant gains in downstream decision quality over the conventional two-stage method. See Figure \ref{fig:LtOF_diagram} (a) and (b) for an illustrative comparison, where the constraint set is denoted with $\cF$.  An overview of related work on the topic is reported in Appendix \ref{app:related_work}.

\subsection*{Challenges in End-to-End Predict-Then-Optimize} 
Despite their advantages over the two-stage, EPO methods face two key challenges:
\textbf{(1)} \textbf{Differentiability}: the need for handcrafted backpropagation rules through $\bm{x}^{\star}(\bm{\zeta})$, which are highly dependent on the form of problem (\ref{eq:opt_generic}), and rely on the assumption of derivatives $ \frac{\partial \bm{x}^{\star}}{\partial \bm{\zeta}}$ which may not exist or provide useful descent directions, and require that the mapping (\ref{eq:opt_generic}) is unique, producing a well-defined function; 
\textbf{(2)} \textbf{Efficiency}: the need to solve the optimization (\ref{eq:opt_generic}) to produce $\bm{x}^{\star}(\bm{\zeta})$ for each sample, at each iteration of training, which is often inefficient even for simple optimization problems. 

This paper is motivated by a need to address these disadvantages. To do so, it recognizes a body of work on training DNNs as \emph{learned optimization proxies} which have fast execution, are automatically differentiable by design, and specialize in learning mappings $\bm{\zeta} \rightarrow \bm{x}^{\star}(\bm{\zeta})$ of hard optimization problems. 
While the next section discusses why the direct application of learned proxies as differentiable optimization solvers in an EPO approach tends to fail, Section \ref{sec:LtOF} presents a successful adaptation of the approach in which optimal solutions are learned end-to-end from the observable features $\bm{z}$.

\section{EPO with Optimization Proxies}
\label{sec:EPOwithProxy}
The Learning-to-Optimize problem setting encompasses a variety of distinct methodologies with the common goal of learning to solve optimization problems. This section characterizes that setting, before proceeding to describe an adaptation of LtO methods to the  Predict-Then-Optimize setting.

\textbf{Learning to Optimize.}
The idea of training DNN models to emulate optimization solvers is referred to as \emph{Learning-to-Optimize (LtO)}  \citep{kotary2021end}. 
Here the goal is to learn a mapping $\bf{F}_\omega: \cC  \to \cX$ 
from the parameters $\bm{\zeta}$ of an optimization problem (\ref{eq:opt_generic}) to its corresponding optimal solution
$\bm{x}^\star(\bm{\zeta})$ (see Figure \ref{fig:LtOF_diagram} (c)). The resulting \emph{proxy} optimization model has as its learnable component a DNN denoted $\hat{\bm{F}}_{\omega}$, which may be augmented with further operations $\bm{\cS}$ such as constraint corrections or unrolled solver steps, so that $\bf{F}_\omega = \bm{\cS} \circ \hat{\bm{F}}_{\omega}$. While training such a lightweight model to emulate optimization solvers is in general difficult, it is made tractable by restricting the task over a \emph{limited distribution} $\Omega^F$ of problem parameters $\bm{\zeta}$.  

\vspace{-2pt}
A variety of LtO methods have been proposed, many of which specialize in learning to solve problems of a specific form. Some are based on supervised learning, in which case precomputed solutions $\bm{x}^{\star}(\bm{\zeta})$ are required as target data in addition to parameters $\bm{\zeta}$ for each sample. Others are \emph{self-supervised}, requiring only knowledge of the problem form (\ref{eq:opt_generic}) along with instances of the parameters $\bm{\zeta}$ for supervision in training. 
LtO methods employ special learning objectives to train the proxy model $\bm{F}_\omega$: 
\begin{equation}\label{eq:LtO_goal}
    \minimize_{\omega}\; \EE_{\bm{\zeta} \sim \Omega^F}
    \left[ \ell^{\text{LtO}} 
        \Big( 
            \bm{F}_\omega(\bm{\zeta}), \bm{\zeta}
        \Big) 
    \right],
\end{equation}
where $\ell^{\text{LtO}}$ represents a loss 
that is specific to the LtO method employed.
A primary challenge in LtO is ensuring the satisfaction of constraints $\bm{g}(\hat{\bm{x}})\leq 0$ and $\bm{h}(\hat{\bm x}) = 0$  by the solutions $\hat{\bm{x}}$ of the proxy model $\bm{F}_\omega$. This can be achieved, exactly or approximately, by a variety of methods, for example iteratively retraining Equation~(\ref{eq:LtO_goal}) while applying dual optimization steps to a Lagrangian loss function \citep{fioretto2020lagrangian,park2023self}, or designing $\bm{\cS}$ to restore feasibility \citep{donti2021dc3}, as reviewed in Appendix \ref{app:lto_methods}. In cases where small constraint violations remain in the solutions $\hat{ \bf{x} }$ at inference time, they can be removed by post-processing with efficient projection or correction methods as deemed suitable for the particular application \citep{kotary2021end}.

\subsection*{EPO with Pretrained Optimization Proxies}
\label{sec:Proxy_PtO}

Viewed from the Predict-then-Optimize lens, learned optimization proxies have two beneficial features by design: {\bf (1)} they enable very fast solving times compared to conventional solvers, and {\bf (2)} are differentiable by virtue of being trained end-to-end. Thus, a natural question is  whether it is possible to use a pre-trained optimization proxy to substitute the differentiable optimization component of an EPO pipeline. Such an approach modifies the EPO objective (\ref{eq:epo_goal}) as:
\begin{equation}\label{eq:LtO_pretrained}
    \!\!\!\!\minimize_\theta\;
    \EE_{(\bm z, \bm{\zeta}) \sim \Omega} 
    \biggl[
    f\Bigl(
        \overbrace{ \pbox{1pt}{gray!20}{\bm{F}_\omega}
         \bigl( \underbrace{\bm{C}_\theta(\bm{z})}_{\hat{\bm{\zeta}}}}^{\hat{\bm x}}\bigr),
        \bm{\zeta}
    \Bigr)
    \biggr],
\end{equation}
\vspace{-2pt}
\!\!in which the solver output $\bm{x}^\star(\hat{\bm{\zeta}})$ of problem (\ref{eq:epo_goal}) is replaced with the prediction $\hat{\bm{x}}$ obtained by LtO model $\bm{F}_\omega$ on input $\hat{\bm{\zeta}}$ (gray color highlights that the model is \!\pbox{0pt}{gray!20}{\text{pretrained}}\!, before freezing its weights $\omega$).

However, a fundamental challenge in LtO lies in the inherent limitation that ML models act as reliable optimization proxies \emph{only within the distribution of inputs they are trained on}. This challenges the implementation of the idea of using pretrained LtOs as components of an end-to-end Predict-Then-Optimize model  as the weights \( \theta \) update during training, leading to continuously evolving inputs \( \bm{C}_{\theta}(\bm{z}) \) to the pretrained optimizer \pbox{1pt}{gray!20}{\bm{F}_{\omega}}. Thus, to ensure robust performance, \pbox{1pt}{gray!20}{\bm{F}_{\omega}} must generalize well across virtually any input during training. However, due to the dynamic nature of \( \theta \), there is an inevitable \emph{distribution shift} in the inputs to \pbox{1pt}{gray!20}{\bm{F}_{\omega}}, destabilizing the EPO training.

\begin{figure}[t]
    \centering
    \includegraphics[width=0.75\linewidth]{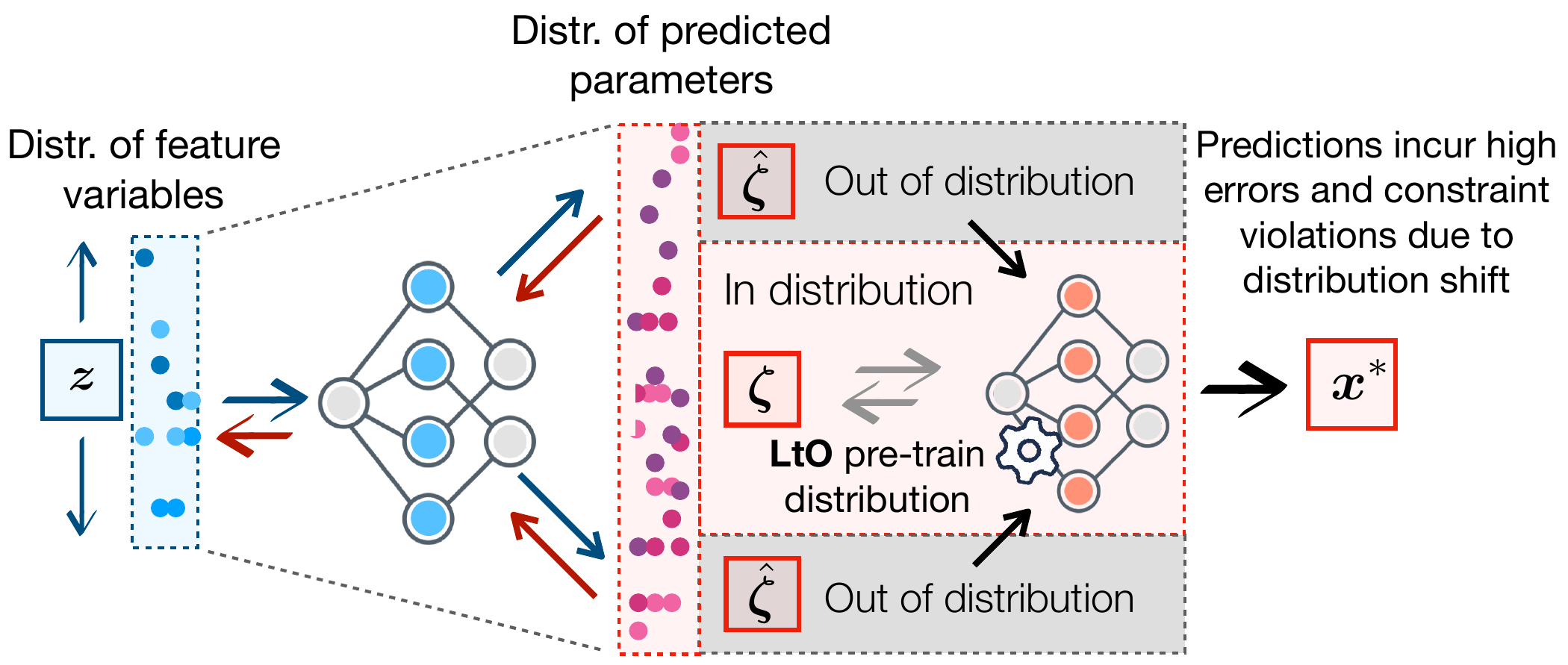}\\
    \caption{A distribution shift between the training distribution of a LtO proxy and the parameter predictions during training leads to inaccuracies in the proxy solver.}
    \label{fig:main-scheme}
\end{figure}

\begin{wrapfigure}[11]{r}{0.48\linewidth}
    \vspace{-10pt}
    \centering
    \includegraphics[width=\linewidth]{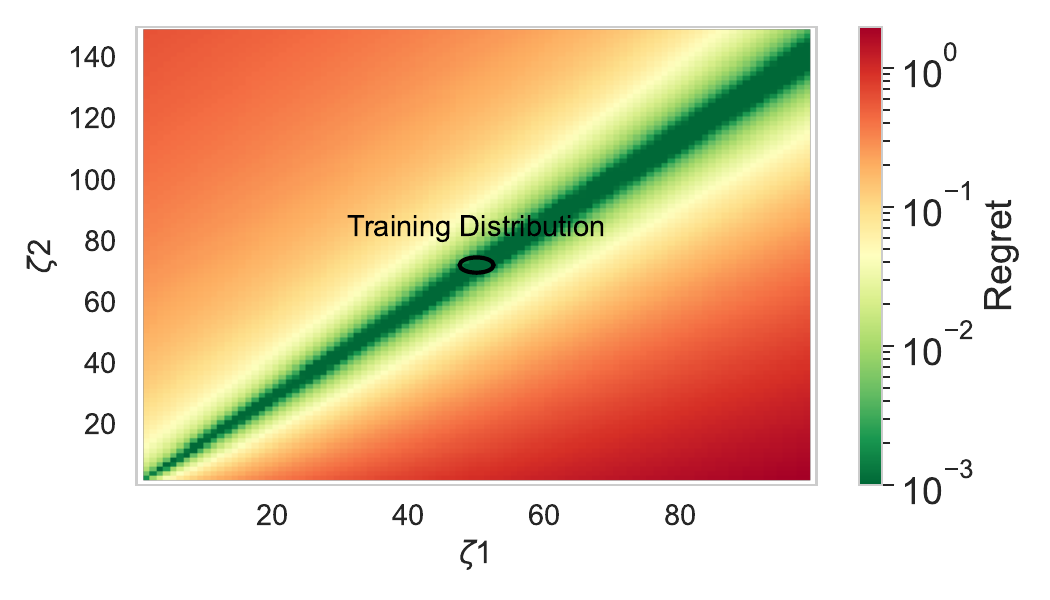}
    \vspace{-24pt}
    \caption{\small Effect on regret as LtO proxy acts outside its training distribution.}
    \label{fig:distribution_shift}
\end{wrapfigure}

Figures \ref{fig:main-scheme} and \ref{fig:distribution_shift} illustrate this issue. The former highlights how the input distribution to a pretrained proxy drifts during EPO training, adversely affecting both output and backpropagation. The latter
quantifies this behavior, exemplified on a simple two-dimensional problem (described in Appendix \ref{app:optimization_problems}), 
showing rapid increase in proxy regret as \( \hat{\bm{\zeta}} \) diverges from the initial training distribution \( \bm{\zeta} \sim \Omega^F \) (shown in black). 
The experimental results presented in Tables \ref{table:portfolio_regret_violation_table},\ref{table:noconvexqp_regret_violation_table}, and \ref{table:acopf_regret_violation_table} reinforce these observations. While each proxy solver performs well within its training distribution, their effectiveness deteriorates sharply when utilized as described in ~\eqref{eq:LtO_pretrained}. This degradation is observed irrespective of any normalization applied to the proxy's input parameters during EPO training.

A step toward resolving this distribution shift issue allows the weights of ${\bm{F}_\omega}$ to adapt to its changing inputs, by \emph{jointly} training the prediction and optimization models:
\begin{equation}\label{eq:joint_dfl_goal}
    \minimize_{\theta, \omega} \;
    \EE_{(\bm z, \bm{\zeta}) \sim \Omega} 
    \biggl[
    f\Bigl(
        \overbrace{ {\bm{F}_\omega}
         \bigl( \underbrace{\bm{C}_\theta(\bm{z})}_{\hat{\bm{\zeta}}}}^{\hat{\bm x}}\bigr), 
        \bm{\zeta}
    \Bigr)
    \biggr].
\end{equation}
The predictive model $\bm{C}_\theta$ is then effectively absorbed into the predictive component of ${\bm{F}_\omega}$, resulting in a \emph{joint} prediction and optimization proxy model $\bm{J}_\phi = {\bm{F}_\omega} \circ \bm{C}_\theta$, where $\phi = (\omega, \theta)$.  Given the requirement for feasible solutions, the training objective (\ref{eq:joint_dfl_goal}) must be replaced with an LtO procedure that enforces the constraints on its outputs. This leads us to the framework presented next. 

\section{Learning to Optimize from Features}
\label{sec:LtOF}

The distribution shift effect described above arises due to the disconnect in training between the first-stage prediction network $\bm{C}_\theta : \cZ \rightarrow \cC$ and the second-stage optimization proxy $\bm{F}_\omega : \cC \rightarrow \cX$. 
However, the Predict-Then-Optimize setting (see Section \ref{sec:PtO}) ultimately only requires the combined model to produce a candidate optimal solution $\hat{\bm{x}} \in \cX$ given an observation of features $\bm{z} \in \cZ$. 
Thus, the intermediate prediction $\hat{\bm{\zeta}} = \bm{C}_\theta(\bm{z})$ in Equation~(\ref{eq:joint_dfl_goal}) is, in principle, not needed.  
This motivates the choice to learn direct mappings from features to optimal solutions of the second-stage decision problem. 
The joint model $\bm{J}_{\phi} : \cZ \to \cX$ is trained by Learning-to-Optimize procedures, employing
\begin{equation}
    \label{eq:LtOF_goal}
    \minimize_{\phi}\;
        \EE_{(\bm{z}, \bm{\zeta}) \sim \Omega} \biggl[
    \ell^{\text{LtO}}\Bigl( 
    \bm{J}_\phi(\bm{z}), \bm{\zeta} \Bigr)
    \biggr].
\end{equation}
This method can be seen as a direct adaptation of the Learn-to-Optimize framework to the Predict-then-Optimize setting. The key difference from the typical LtO setting, described in Section \ref{sec:EPOwithProxy}, is that problem parameters $\bm{\zeta} \in \cC$ are not known as inputs to the model, but the correlated features $\bm{z} \in \cZ$ are known instead. Therefore, estimated optimal solutions now take the form $\hat{\bm{x}} = \bm{J}_\phi(\bm{z})$ rather than $\hat{\bm{x}} =  \bm{F}_\omega(\bm{\zeta})$. Notably, this causes the self-supervised LtO methods to become \emph{supervised}, since the ground-truth parameters $\bm{\zeta} \in \cC$ now act only as target data while the separate feature variable  $\bm{z}$ takes the role of input data.

We refer to this approach as \emph{Learning to Optimize from Features (LtOF)}. Figure \ref{fig:LtOF_diagram} illustrates the key distinctions of LtOF relative to the other learning paradigms studied in the paper. Figures (\ref{fig:LtOF_diagram}c) and (\ref{fig:LtOF_diagram}d) distinguish LtO from LtoF by a change in model's input space, from $\bm{\zeta} \in \cC$ to $\bm{z} \in \cZ$. This brings the framework into the same problem setting as that of the two-stage and end-to-end PtO approaches, illustrated in Figures (\ref{fig:LtOF_diagram}a) and (\ref{fig:LtOF_diagram}b). The key difference from the PtO approaches is that they produce an estimated optimal solution $\bm{x}^{\star}(\hat{\bm{\zeta}})$  by using a true optimization solver, but applied to an imperfect parametric prediction $\hat{\bm{\zeta}} = \bm{C}_\theta(\bm z)$. 
In contrast, LtOF directly estimates optimal solution $\hat{ \bm{x} }(\bm{z}) = \bm{J}_\phi(\bm{z})$ from features $\bm{z}$, circumventing the need to represent an estimate of $\bm{\zeta}$. 

\subsection{Sources of Error}
Both the PtO and LtOF methods yield solutions subject to \( \emph{regret} \), which measures suboptimality relative to the true parameters \( \bm{\zeta} \), as defined in Equation \ref{eq:regret_def}. However, while in end-to-end and, especially, in the two-stage PtO approaches, the regret in $\bm{x}^{\star}(\hat{\bm{\zeta}})$ arises from imprecise parameter predictions $\hat{\bm{\zeta}} = \bm{C}_{\theta}(\bm{z})$ \citep{mandi2023decision}, 
in LtOF, the regret in the inferred solutions $\hat{\bm{x}}(\bm{z}) = \bm{J}_\phi(\bm{z})$ arises due to imperfect learning of the proxy optimization. This error is inherent to the LtO methodology used to train the joint prediction and optimization model $\bm{J}_{\phi}$, and persists even in typical LtO, in which $\bm{\zeta}$ are precisely known. In principle, a secondary source of error can arise from imperfect learning of the implicit feature-to-parameter mapping $\bm{z} \to \bm{\zeta}$ within the joint model $\bm{J}_{\phi}$. However, these two sources of error are indistinguishable, as the prediction and optimization steps are learned jointly. Finally, depending on the specific LtO procedure adopted, a further source of error arises when small violations to the constraints occur in $\hat{\bm{x}}(\bm{z})$. In such cases, restoring feasibility (e.g, through projection or heuristics steps) 
often induces slight increases in regret \citep{fioretto2020lagrangian}. 

Despite being prone to optimization error, Section \ref{sec:Experiments} shows that Learning to Optimize from Features greatly outperforms two-stage methods, and is competitive with EPO training based on exact differentiation through $\bm{x}^{\star}(\bm{\zeta})$, when the feature-to-parameter mapping $\bm{z} \to \bm{\zeta}$ is complex. This is achieved \emph{without} any access to exact optimization solvers, nor models of their derivatives.  This feat can be explained by the fact that by learning optimal solutions end-to-end directly from features, LtOF does not directly depend on learning an accurate representation of the underlying mapping from $\bm{z}$ to $\bm{\zeta}$.

\subsection{Efficiency Benefits} 
Because the primary goal of the Learn-to-Optimize methodology is to achieve \emph{fast solving times}, the LtOF approach broadly inherits this advantage. While these benefits in speed may be diminished when constraint violations are present and complex feasibility restoration are required, 
efficient feasibility restoration is possible for many classes of optimization models \cite{beck2017first}. This enables the design of \emph{accelerated} PtO models within the LtOF framework, as shown in Section \ref{sec:Experiments}.

\subsection{Modeling Benefits} While EPO approaches require the implementation of problem-specific backpropagation rules, the LtOF framework allows for the utilization of existing LtO methodologies in the PtO setting, on a problem-specific basis. A variety of existing LtO methods specialize in learning to solve convex and nonconvex optimization \citep{fioretto2020lagrangian,park2023self,donti2021dc3}, combinatorial optimization \citep{bello2017neural,kool2019attention}, and other more specialized problem forms \citep{wu2022deep}. The experiments of this paper focus on the scope of continuous optimization problems, whose LtO approaches share a common set of solution strategies.

\section{Experiments}
\label{sec:Experiments}

This section evaluates three distinct LtO methods adapted to the LtOF setting, on three different Predict-Then-Optimize tasks, where each task involves a distinct second stage optimization component $\bm{x}^{\star}: \cC \rightarrow \cX$, as in \eqref{eq:opt_generic}. 
These include a convex quadratic program (QP), a nonconvex quadratic programming variant, and a nonconvex AC-Optimal Power Flow problem, to demonstrate the general utility of the framework. First, the section's three LtOF methods are briefly described.

\subsection{Learning to Optimize Methods}
\label{app:lto_methods}

This section reviews in more depth those LtO methods which are adapted to solve PtO problems in Section \ref{sec:Experiments} of this paper.
Each description below assumes a DNN model $\hat{\bm{F}}_{\omega}$, which acts on parameters $\bm{\zeta_i}$ specifying an instance of problem \eqref{eq:opt_generic}, to produce an estimate of the optimal solution  $\hat{\bm{x}}\coloneqq \bm{F}_{\omega}(\bm{\zeta}) $, so that  $\hat{\bm{x}} \approx \bm{x}^{\star}(\bm{\zeta})$.

\subsubsection{Lagrangian Dual Learning (LD)} 
\citet{fioretto2020lagrangian} uses the following modified Lagrangian loss function for training  $\hat{\bm{x}} = \bm{F}_{\omega}(\bm{\zeta})$:
\begin{equation}
    \label{eq:ld_loss}
    \cL_{\textbf{LD}}(\hat{\bm{x}},\bm{\zeta}) =
        \| \hat{\bm{x}} - \bm{x}^{\star}(\bm{\zeta}) \|^2_2 +  \bm{\lambda}^T \left[ \bm{g}(\hat{\bm{x}},\bm{\zeta}) \right]_+ + \bm{\mu}^T  \bm{h}(\hat{\bm{x}},\bm{\zeta}).
\end{equation}

At each iteration of LD training, the model $\bm{F}_{\omega}$ is trained to minimize the loss $\cL_{\textbf{LD}}$. Then, updates to the multiplier vectors $\lambda$ and $\mu$ are calculated based on the average constraint violations incurred by the predictions $\hat{\bm{x}}$, mimicking a dual ascent method \cite{boyd2011distributed}. In this way, the method minimizes a balance of constraint violations and proximity to the precomputed target optima $\bm{x}^{\star}(\bm{\zeta})$.

\subsubsection{Self-Supervised Primal-Dual Learning (PDL)} 
\citet{park2023self} use an augmented Lagrangian loss for self-supervised learning:
\begin{equation}
    \label{eq:pdl_loss}
    \cL_{\textbf{PDL}}(\hat{\bm{x}},\bm{\zeta}) =
         f(\hat{\bm{x}}, \bm{\zeta}) + \hat{\bm{\lambda}}^T  \bm{g}(\hat{\bm{x}},\bm{\zeta})  + \hat{\bm{ \mu }}^T \bm{h}(\hat{\bm{x}},\bm{\zeta}) + \frac{\rho}{2} \left( \sum_j \nu(g_j(\hat{\bm{x}})) + \sum_j \nu(h_j(\hat{\bm{x}}   )) \right),
\end{equation}
where $\nu$ measures the constraint violation. At each iteration of PDL training, a separate estimate of the Lagrange multipliers is stored for each problem instance in training, and updated by an augmented Lagrangian method \cite{boyd2011distributed} after training  $\hat{\bm{x}} = \bm{F}_{\omega}(\bm{\zeta}) $ to minimize \eqref{eq:pdl_loss}. In addition to the primal network $\bm{F}_{\omega}$, a dual network $\cD_{\bm{\omega'}}$ learns to store updates of the multipliers for each instance, and predict them as $(\hat{\bm{\lambda}}, \hat{\bm{\mu}}) = \cD_{\bm{\omega'}}(\bm{\zeta})$ to the next iteration.

\subsubsection{Deep Constraint Completion and Correction (DC3)} 
\citet{donti2021dc3} use the loss function
\begin{equation}
    \label{eq:dc3_loss}
    \cL_{\textbf{DC3}}(\hat{\bm{x}},\bm{\zeta}) =
         f(\hat{\bm{x}}, \bm{\zeta}) + \lambda \| \left[ \bm{g}(\hat{\bm{x}},\bm{\zeta}) \right]_+ \|_2^2 +  \mu  \| \bm{h}(\hat{\bm{x}},\bm{\zeta}) \|_2^2
\end{equation}
which combines a problem's objective value with two additional terms which aggregate the total violations of its equality and inequality constraints. The scalar multipliers $\lambda$ and $\mu$ are not adjusted during training. However, feasibility of predicted solutions is enforced by treating $\hat{\bm{x}} =  \hat{\bm{F}}_{\omega}(\bm{\zeta})$ as an estimate for only a subset of optimization variables. The remaining variables are completed by solving the underdetermined equality constraints $\bm{h}(\bm{\hat{x}})=\bm{0}$ as a system of equations. Inequality violations are corrected by gradient descent on the their aggregated values  $\| \left[ \bm{g}(\hat{\bm{x}},\bm{\zeta}) \right]_+ \|^2$ . These completion and correction steps form the function $\bm{S}$, where $\bm{F}_{\omega}(\bm{\zeta}) = \bm{S} \circ \hat{\bm{F}}_{\omega}(\bm{\zeta})$.

While several other Learning-to-Optimize methods have been proposed in the literature, the above-described collection represents diverse subset which is used to demonstrate the potential of adapting the end-to-end LtO methodology as a whole to the Predict-Then-Optimize setting. 

\subsection{Experimental Settings}
\textbf{Feature generation}. End-to-End Predict-Then-Optimize methods integrate learning and optimization 
to minimize the propagation of prediction errors--specifically, from feature mappings \( \bm{z} \to \bm{\zeta} \) to the resulting decisions \( \bm{x}^{\star}(\bm{\zeta}) \) (regret). It's crucial to recognize that \emph{even methods with high error propagation} can yield low regret \emph{if the prediction errors are low}. To account for this, EPO studies often employ synthetically generated feature mappings to control prediction task difficulty \citep{elmachtoub2020smart,mandi2023decision}. 
Accordingly, for each experiment, we generate feature datasets \( (\bm{z}_1, \ldots \bm{z}_N) \in \mathcal{Z} \) from ground-truth parameter sets \( (\bm{\zeta}_1, \ldots \bm{\zeta}_N) \in \mathcal{C} \) using random mappings of increasing complexity. A feedforward neural network, \( \bm{G}^k \), initialized uniformly at random with \( k \) layers, serves as the feature generator \( \bm{z} = \bm{G}^k(\zeta) \). Evaluation is then carried out for each PtO task on feature datasets generated with \( k \in \{ 1,2,4,8\} \), keeping target parameters $\bm{\zeta}$ constant.

\textbf{Baselines}. In our experiments, LtOF models use feedforward networks with \( k \) hidden layers. For comparison, we also evaluate two-stage and, where applicable, EPO models, using architectures with \( k \) hidden layers where \( k \in \{ 1,2,4,8 \} \). Further training specifics are provided in Appendix \ref{app:experimental}.

\textbf{Comparison to LtO setting}. It is natural to ask how solution quality varies when transitioning from LtO to LtOF in a PtO setting, where solutions are learned directly from features. To address this question, each PtO experiment includes results from its analogous Learning to Optimize setting, where a DNN  $\bf{F}_\omega : \cC \to \cX$ learns a mapping from the parameters $\bm \zeta$ of an optimization problem to its corresponding solution $\bm{x}^\star(\bm{\zeta})$. This is denoted \( k\!=\!0 \; (\text{LtO}) \), indicating the absence of any feature mapping. 
All figures report the regret obtained by LtO methods for reference, although they are not directly comparable to the Predict-then-Optimize setting.

\textbf{Comparison to EPO with Pretrained Proxy}. The end-to-end LtOF implementations are also compared against EPO models with pre-trained optimization proxies as a baseline, as described in Section \ref{sec:EPOwithProxy}. 

All reported results are averages across 20 random seeds and the reader is referred to Appendix \ref{app:experimental} for additional details regarding experimental settings, architectures, and hyperparamaters adopted.

\begin{figure}[t]
    \centering
    \begin{minipage}{0.53\textwidth}
        \centering
        \hspace{-12pt}
        \includegraphics[width=\linewidth]{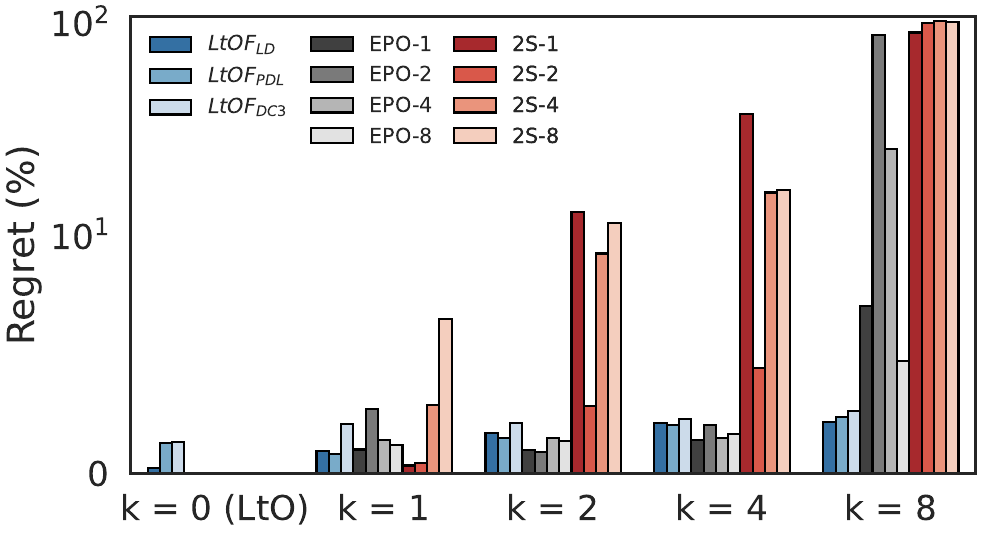}
        \caption{\small Comparison between \textbf{LtO} ($k\!=\!0$), \textbf{LtOF}, Two-stage (\textbf{2S}) and \textbf{EPO} ($k\!>\!1$) on the portfolio optimization. 2S(EPO)-$m$ indicates that the prediction model of the respective PtO method is an $m$ layer ReLU neural network.}
        \label{fig:portfolio_results}
    \end{minipage}
    \hfill
    \begin{minipage}{0.45\textwidth}
        \centering
        \vspace{0pt}
        \resizebox{\textwidth}{!}{%
        \begin{tabular}{clccc}
            \toprule
            & {\bf Method} & {\bf Portfolio} & {\bf N/conv.~QP} & {\bf AC-OPF} \\
            \midrule
            \multirow{6}{*}{\rotatebox{90}{\textbf{LtOF}}}
            & LD it     & {\bf 0.0003} & 0.0000 & {\bf 0.0004} \\
            & \gray{LD fct}    & \gray{0.0000} & \gray{0.0045} & \gray{0.1573} \\
            & PDL it    & {\bf 0.0003} & 0.0000 & 0.0006 \\
            & \gray{PDL fct}   & \gray{0.0000} & \gray{0.0045} & \gray{0.1513} \\
            & DC3 it    & 0.0011       & 0.0001 & - \\
            & \gray{DC3 fct}   & \gray{0.0003}  & \gray{0.0000} & - \\
            \midrule
        \multirow{4}{*}{\rotatebox{90}{\textbf{PtO}}}
            & PtO-1 et  & 0.0054 & 0.0122 & 0.1729 \\
            & PtO-2 et  & 0.0059 & 0.0104 & 0.1645 \\
            & PtO-4 et  & 0.0062 & 0.0123 & 0.1777 \\
            & PtO-8 et  & 0.0067 & 0.0133 & 0.1651 \\
        \bottomrule
        \end{tabular}
        }
        \vspace{0pt}
        \captionof{table}{\small Execution ({\it et}), inference ({\it it}), and feasibility correction ({\it fct}) times for {\bf LtOF} and {\bf PtO} (in seconds) for each sample. {\bf Two-stage} methods execution times are comparable to PtO's ones.}
        \label{table:inference_times}
    \end{minipage}%
\end{figure}

\subsection{Convex Quadratic Portfolio Optimization}
\label{exp:CQO}
A well-known problem combining prediction and optimization is the Markowitz Portfolio Optimization \citep{rubinstein2002markowitz}. This task has as its optimization component a convex Quadratic Program:
\begin{equation} 
    \label{eq:opt_portfolio}
        \bm{x}^{\star}(\bm{\zeta}) =  \argmax_{\bm{x} \geq \bm{0}} \; \bm{\zeta}^T \bm{x} - \lambda \bm{x}^T \bm{\Sigma} \bm{x}, \;\;\; \text{s.t. } \; \bm{1}^T \bm{x} = 1
\end{equation}
in which parameters $\bm{\zeta} \in \mathbb{R}^D$ represent future asset prices, and decisions $\bm{x} \in \mathbb{R}^D$ represent their fractional allocations within a portfolio. The objective is to maximize a balance of risk, as measured by the quadratic form covariance matrix $\Sigma$, and total return $\bm{\zeta}^T \bm{x}$. Historical prices of $D = 50$ assets are obtained from the Nasdaq online database \citep{NASDAQ} and used to form price vectors $\bm{\zeta}_i, \; 1\leq i \leq N$, with $N\!=\!12,000$ individual samples collected from 2015-2019. In the outputs $\hat{\bm{x}}$ of each LtOF method, possible feasibility violations are restored, at \emph{low computational cost}, by first clipping $[\hat{\bm{x}}]_+$ to satisfy $\bm{x} \geq \bm{0}$, then dividing by its sum to satisfy $\bm{1}^T \bm{x} = 1$. The convex solver \texttt{cvxpy} \citep{diamond2016cvxpy} is used as the optimization component in each PtO method.

\textbf{Results}. Figure \ref{fig:portfolio_results} shows the percentage regret due to LtOF implementations based on \textit{LD}, \textit{PDL} and \textit{DC3}. Two-stage and EPO models are evaluated for comparison,  with predictive components given various numbers of layers. For feature complexity $k>1$, each LtOF model outperforms the best two-stage model, increasingly with $k$ and up to nearly \emph{two orders of magnitude} when $k=8$.  
The EPO model, trained using exact derivatives  through  (\ref{eq:opt_portfolio}) as provided by the differentiable solver in \texttt{cvxpylayers} \citep{agrawal2019differentiable} is competitive with LtOF until $k=4$, after which point its best variant is outperformed by each LtOF variant. This result showcases the ability of LtOF models to reach high accuracy under complex feature mappings \emph{without} access to optimization problem solvers \emph{or} their derivatives, in training or inference, in contrast to conventional PtO frameworks. Full accuracy results are reported in Table \ref{table:portfolio_regret_violation_table}, which includes constraint violation and regret of the inferred solutions before feasibility restoration. 

Table \ref{table:inference_times} presents LtOF inference times (\( \textit{it} \)) and feasibility correction times (\( \textit{fct} \)), which are compared with the per-sample execution times (\( \textit{et} \)) for PtO methods. Run times for two-stage methods are  closely aligned with those of EPO, and thus obmitted.
Notice how the LtOF methods are at least an order of magnitude faster than PtO methods. This efficiency has two key implications: firstly, the per-sample speedup can significantly accelerate training for PtO problems. Secondly, it is especially advantageous during inference, particularly if data-driven decisions are needed in real-time.

\begin{table}[t]
\centering
    \resizebox{0.95\textwidth}{!}{%
    \begin{tabular}{clccccc}
    \toprule
    & Method & $k=0 \; (LtO)$ & $k=1$ & $k=2$ & $k=4$ & $k=8$ \\
    \midrule
    \multirow{1}{*}{\rotatebox{90}{LtOF}}
    & LD Regret                 &  \bf{1.2785} &      0.9640  &     1.7170   &     2.1540   &  \bf{2.1700} \\
    & \rem{LD Regret (*)}       & \rem{1.1243} & \rem{1.0028} & \rem{1.5739} & \rem{2.0903} & \rem{2.1386} \\
    & \rem{LD Violation (*)}    & \rem{0.0037} & \rem{0.0023} & \rem{0.0010} & \rem{0.0091} & \rem{0.0044} \\
    & PDL Regret                &      1.2870  &      0.8520  &      1.5150  &      2.0720  &       2.3830 \\
    & \rem{PDL Regret (*)}      & \rem{1.2954} & \rem{0.9823} & \rem{1.4123} & \rem{1.9372} & \rem{2.0435} \\
    & \rem{PDL Violation (*)}   & \rem{0.0018} & \rem{0.0097} & \rem{0.0001} & \rem{0.0003} & \rem{0.0003} \\
    & DC3 Regret                &      1.3580  &      2.1040  &      2.1490  &      2.3140  & 2.6600       \\
    & \rem{DC3 Regret (*)}      & \rem{1.2138} & \rem{1.8656} & \rem{2.0512} & \rem{1.9584} & \rem{2.3465} \\
    & \rem{DC3 Violation (*)}   & \rem{0.0000} & \rem{0.0000} & \rem{0.0000} & \rem{0.0000} & \rem{0.0000} \\
   \midrule
    \multirow{3}{*}{}
    & Two-Stage Regret (Best)   &      -       &  \bf{0.3480} &      2.8590  &      4.4790  &     91.3260 \\
    & EPO Regret (Best)         &      -       &      1.0234  &  \bf{0.9220} &  \bf{1.4393} &      4.7495 \\
    & EPO Proxy Regret (Best)   &      -       &    136.4341  &    154.3960  &    119.3082  &    114.6953 \\
    \bottomrule
    \end{tabular}
    }
    \caption{Regret and Constraint Violations for Portfolio Experiment. (*) denotes ``Before Restoration''.}
    \label{table:portfolio_regret_violation_table}
\end{table}

\subsection{Nonconvex QP Variant}

As a step in difficulty beyond convex QPs, this experiment considers a generic QP problem augmented with an additional oscillating objective term, resulting in a \emph{nonconvex} optimization component:

\begin{subequations}
    \label{eq:bilinear}
    \begin{align*}
        \mathbf{x}^{\star}(\bm{\zeta}) = \argmin_{\bm{x}} &\;\;
        \frac{1}{2} \bm{x}^T \bm{Q}  \bm{x} + \bm{\zeta}^T \sin( \bm{x} ) \\
        \texttt{s.t.} \;\; 
        & \bm{A} \bm{x} = \bm{b}, \; \bm{G} \bm{x} \leq \bm{h},
    \end{align*}
\end{subequations}

in which the $\sin$ function is applied elementwise. This formulation was used to evaluate the LtO methods proposed both in \cite{donti2021dc3} and in \cite{park2023self}. Following those works, $\bm{0} \preccurlyeq \bm{Q} \in \mathbb{R}^{n \times n}$, $\bm{A} \in \mathbb{R}^{n_{\text{eq}} \times n}$, $\bm{b} \in \mathbb{R}^{n_{\text{eq}}}$, $\bm{G} \in \mathbb{R}^{n_{\text{ineq}} \times n}$ and $\bm{h} \in \mathbb{R}^{n_{\text{ineq}}}$ have elements drawn uniformly at random. Here it is evaluated as part of a Predict-Then-Optimize pipeline in which predicted coefficients occupy the nonconvex term. Feasibility is restored by a projection onto the feasible set, which is calculated by a more efficiently solvable \emph{convex} QP. The problem dimensions are $n=50$ $n_{\text{eq}} = 25$, and $n_{\text{ineq}}=25$.
\begin{figure}[t]
        \centering
        \includegraphics[width=0.48\linewidth]{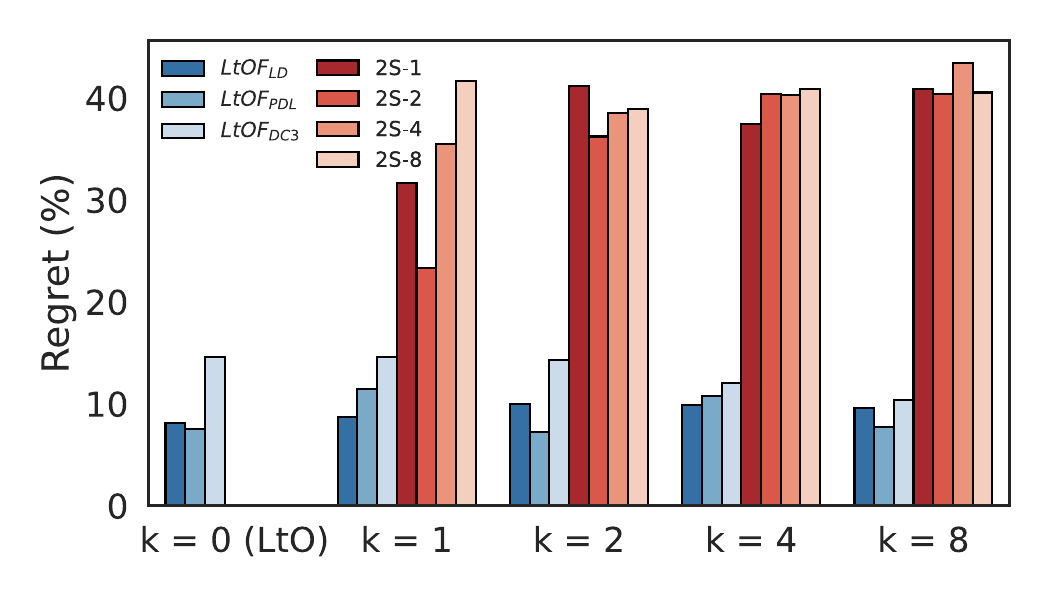}
        \hfill
        \includegraphics[width=0.48\linewidth]{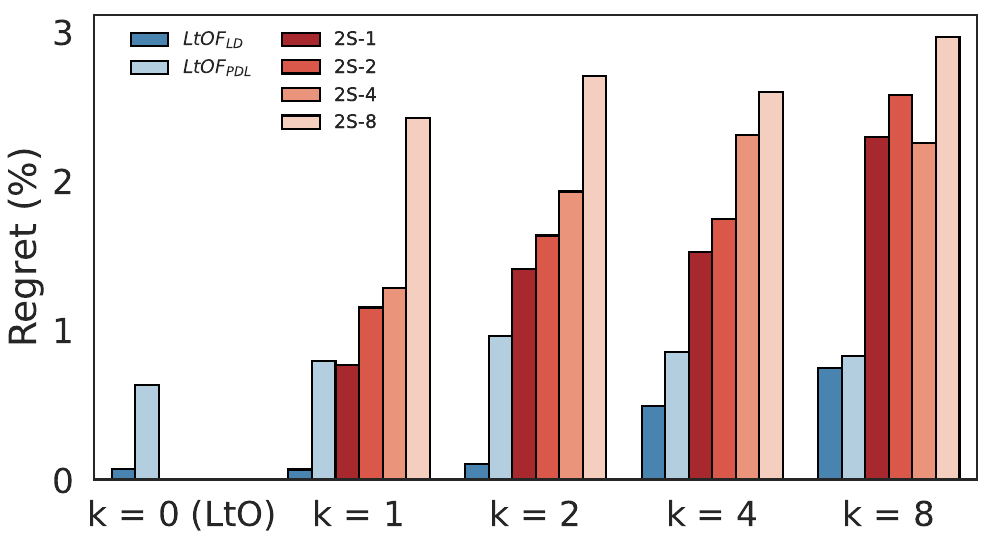}
        \caption{\small Comparison between LtO ($k=0$), LtOF, and Two Stage Method (2S) on the nonconvex QP (\textbf{left}) and 
        AC-OPF case (\textbf{right}). Right plot y-axis is in log-scale.}
        \label{fig:nc_results}
\end{figure}

\textbf{Results.}
Figure \ref{fig:nc_results} (left) shows regret due to LtOF models based on  \textit{LD}, \textit{PDL} and \textit{DC3}, along with two-stage baseline PtO methods. No EPO baselines are available due to the optimization component's nonconvexity. The best two-stage models perform poorly for most values of $k$, implying that the regret is particularly sensitive to prediction errors in the oscillating term. Thus its increasing trend with $k$ is less pronounced than in other experiments. The best LtOF models achieve over $4$ times lower regret than the best baselines, suggesting strong potential for this approach in contexts which require predicting parameters of non-linear objective functions. Additionally, the fastest LtOF method achieves up to three order magnitude 
speedup over the two-stage, after restoring feasibility.

\begin{table}[t]
\centering
\resizebox{0.95\textwidth}{!}{
    \begin{tabular}{clccccc}
    \toprule
        & Method & $k=0 \; (LtO)$ & $k=1$ & $k=2$ & $k=4$ & $k=8$ \\
        \midrule
        \multirow{9}{*}{\rotatebox{90}{LtOF}}
        & LD Regret                 &  8.0757      & \bf{8.6826}  &     9.9279   &  \bf{9.7879} & 9.5473       \\
        & \rem{LD Regret (*)}       & \rem{8.1120} & \rem{8.7416} & \rem{9.9250} & \rem{9.8211} & \rem{9.5556} \\
        & \rem{LD Violation (*)}    & \rem{0.0753} & \rem{0.0375} & \rem{0.0148} & \rem{0.0162} & \rem{0.0195} \\
        & PDL Regret                &  \bf{7.4936} &      11.424  &  \bf{7.2699} &    10.7474   & \bf{7.6399}  \\
        & \rem{PDL Regret (*)}      & \rem{7.7985} & \rem{11.429} & \rem{7.2735} & \rem{10.749} & \rem{7.6394} \\
        & \rem{PDL Violation (*)}   & \rem{0.0047} & \rem{0.0032} & \rem{0.0028} & \rem{0.0013} & \rem{0.0015} \\
        & DC3 Regret                &   13.946    &   14.623     & 14.271      &  11.028      & 10.666      \\
        & \rem{DC3 Regret (*)}      & \rem{14.551} & \rem{14.517} & \rem{13.779} & \rem{11.755} & \rem{10.849} \\
        & \rem{DC3 Violation (*)}   & \rem{1.4196} & \rem{0.8259} & \rem{0.5158} & \rem{0.5113} & \rem{0.5192} \\
       \midrule
       \multirow{2}{*}{}
        & Two-Stage Regret (Best)   & -            &      23.2417 &      36.1684 &     37.3995  &    38.2973\\
        & EPO Proxy Regret (Best)   & -            &      793.2369 &      812.7521 &     804.2640  &    789.5043  \\
    \bottomrule
    \end{tabular}
    }
\caption{Regret and Constraint Violations for Nonconvex QP Experiment. (*) denotes ``Before Restoration''.}
\label{table:noconvexqp_regret_violation_table}
\end{table}

\subsection{Nonconvex AC-Optimal Power Flow}

Given a vector of marginal costs $\bm \zeta$ for each power generator in an electrical grid, the AC-Optimal Power Flow problem optimizes the generation and dispatch of electrical power from generators to nodes with predefined demands. The objective is to minimize cost, while meeting demand exactly. The full optimization problem and more details are specified in Appendix \ref{app:optimization_problems}, where a quadratic cost objective is minimized subject to nonconvex physical and engineering power systems constraints. This experiment simulates a energy market situation in which generation costs are as-yet unknown to the power system planners, and must be estimated based on correlated data. The overall goal is to predict costs so as to minimize cost-regret over an example network with $54$ generators, $99$ demand loads, and $118$ buses taken from the well-known NESTA energy system test case archive \citep{coffrin2014nesta}. Feasibility is restored for each LtOF model by a projection onto the nonconvex feasible set. Optimal solutions to the AC-OPF problem, along with such projections, are obtained using state-of-the-art Interior Point OPTimizer IPOPT \citep{ipopt}.

\begin{table}[t]
\centering
    \resizebox{0.95\textwidth}{!}{%
    \begin{tabular}{clccccc}
    \toprule
    & Method & $k=0 \; (LtO)$ & $k=1$ & $k=2$ & $k=4$ & $k=8$ \\
    \midrule
    \multirow{6}{*}{\rotatebox{90}{LtOF}}
    & LD Regret                 &      0.0680  &     {\bf 0.0673}  & {\bf 0.1016}&  {\bf 0.4904} & {\bf 0.7470} \\
    & \rem{LD Regret (*)}       & \rem{0.0009} & \rem{0.0009} & \rem{0.0013} & \rem{0.0071} & \rem{0.0195} \\
    & \rem{LD Violation (*)}    & \rem{0.0035} & \rem{0.0017} & \rem{0.0020} & \rem{0.0037} & \rem{0.0042} \\
    & PDL Regret                &      0.6305  &      0.7958  &      0.9603  &      0.8543  &      0.8304 \\
    & \rem{PDL Regret (*)}      & \rem{0.0210} & \rem{0.0242} & \rem{0.0260} & \rem{0.0243} & \rem{0.0242} \\
    & \rem{PDL Violation (*)}   & \rem{0.0001} & \rem{0.0002} & \rem{0.0000} & \rem{0.0002} & \rem{0.0002} \\
    \midrule
    \multirow{2}{*}{}
    & Two-Stage Regret (Best)   & -            & 0.7620 &      1.4090  &      1.5280 &       2.4740 \\
    & EPO Proxy Regret (Best)  & -            & 431.7664 &      389.0421  &      413.8941 &       404.7452 \\
    \bottomrule
    \end{tabular}
    }
    \caption{Regret and Constraint Violations for AC-OPF Experiment. (*) denotes ``Before Restoration''.}
    \label{table:acopf_regret_violation_table}
\end{table}

\section{Related Work}
\label{app:related_work}

\paragraph{Predict-Then-Optimize}
While the idea is general and has broader applications, differentiation through the optimization of \eqref{eq:opt_generic} is central to EPO approaches.   Differentiation of quadratic programming problems was introduced by \cite{amos2019optnet}, which  implicitly differentiates the solution via its KKT equations of optimality, proposes its use for defining general-purpose learnable layers in neural networks. \citet{agrawal2019differentiating} proposes a more general differentiable cone programming solver, which is leveraged by \cite{agrawal2019differentiable} to solve and differentiate general convex programs, by pairing it with a symbolic system for conversion of convex programs to canonical cone programs. \citet{kotary2023folded} shows how to differentiate optimization problems by leveraging automatic differentiation through a single step of a convergent solution method to implicitly differentiate its fixed-point conditions.
For discrete problems such as linear programs, the mapping defined by \eqref{eq:opt_generic} is piecewise constant and cannot be differentiated. \citep{elmachtoub2020smart} propose a surrogate loss function for \eqref{eq:epo_goal} in cases where $f$ is linear, which admits useful subgradients. \citep{wilder2018melding}  proposes backpropagation through linear programs by adding a smooth quadratic term to the objective and differentiating the resulting QP problem via \cite{amos2019optnet}, and \cite{ferber2020mipaal} extends the technique to mixed-integer programs via the equivalent linear program found by cutting planes. \citep{berthet2020learning} also propose backpropagation through linear programs but by smoothing the mapping \eqref{eq:opt_generic} through  random noise perturbations to the objective function. \citep{vlastelica2020differentiation} form approximate derivatives through linear optimization of discrete variables, by using finite difference approximations.

\paragraph{Learning to Optimize}
The comprehensive survey \cite{bengio2021machine} focuses on machine learning methods aimed at boosting combinatorial solvers by predicted intermediate information. Related works involve learning heurstics for combinatorial solvers including branching rules \cite{khalil2016learning} and cutting rules \cite{deza2023machine} in conventional mixed-integer programming.
For continuous problems, learning of active constraints \cite{misra2022learning} and learning warm-starts \cite{sambharya2023end} are two ways in which intermediate information can be learned to accelerate optimization solvers on a problem instance-specific basis. However, such methods are not relevant to the idea of learning to optimize from features, since they do not produce solutions end-to-end from parameters, but rather intermediate information utilized by an offline solver. End-to-end learning for combinatorial optimization appeared as early as \cite{vinyals2017pointer}, followed by \cite{bello2017neural} which extended the idea to an unsupervised setting by training with reinforcement learning with policy gradient methods. The policy gradient method has been adapted to combinatorial problems such as vehicle routing \cite{kool2018attention} and job scheduling \cite{mao2019learning}, and generally relies on softmax representations of permutations and subset selections to enforce feasibility. Learning combinatorial solutions via supervised penalty methods was proposed in \cite{kotary2022fast, kotary2021learning}. General frameworks for end-to-end learning of non-combinatorial problems have been proposed in the works \cite{fioretto2020predicting,park2023self,donti2021dc3}, which are each reviewed and incorporated in the experiments of Section \ref{sec:Experiments}.

\section{Limitations, Discussion, and Conclusions}
The primary \emph{advantage} of the Learning to Optimize from Features approach to PtO settings is its generic framework, which enables it to leverage a variety of existing techniques and methods from the LtO literature. On the other hand, as such, a particular implementation of LtOF may inherit any limitations of the specific LtO method that it adopts. For example, when the LtO method does not ensure feasibility, the ability to restore feasibility may be need as part of a PtO pipeline. Future work should focus on understanding to what extent a broader variety of LtO methods can be applied to PtO settings; given the large variety of existing works in the area, such a task is beyond the scope of this paper. 
In particular, this paper does not investigate of the use of \emph{combinatorial} optimization proxies in learning to optimize from features. Such methods tend to use a distinct set of approaches from those studied in this paper, often relying on training by reinforcement learning \citep{bello2017neural,kool2019attention,mao2019learning}, and are not suited for capturing broad classes of optimization problems. As such, this direction is left to future work. 

The main \emph{disadvantage} inherent to any LtOF implementation, compared to end-to-end PtO, is the inability to recover parameter estimations from the predictive model, since optimal solutions are predicted end-to-end from features. Although it is not required in the canonical PtO problem setting, this may present a complication in situations where transferring the parameter estimations to external solvers is desirable. This presents an interesting direction for future work.

By showing that effective Predict-Then-Optimize models can be composed purely of Learning-to-Optimize methods, this paper has aimed to provide a unifying perspective on these as-yet distinct problem settings. The flexibility of its approach has been demonstrated by showing superior performance over PtO baselines with diverse problem forms. As the advantages of LtO are often best realized in combination with application-specific techniques, it is hoped that future work can build on these findings to maximize the practical benefits offered by Learning to Optimize in settings that require data-driven decision-making.

\section{Acknowledgement}
This research was partially supported by NSF grants 2232054, 2242931, and 2143706. Its views and conclusions are to be considered as of the authors only. 

\bibliography{iclr2024_conference}
\bibliographystyle{iclr2024_conference}

\appendix


\section{Optimization Problems}
\label{app:optimization_problems}

\paragraph{Illustrative $2D$ example}
Used for illustration purposes, the $2D$ optimization problem used to produce the results of Figure \ref{fig:distribution_shift} takes the form
\begin{subequations}
    \label{eq:dist_shift_fig_dummy_prob}
    \begin{align*}
        \mathbf{x}^{\star}(\bm{\zeta}) = \argmin_{\bm{x} } &\;\;
        \zeta_1 x_1^2 + \zeta_2 x_2^2\\
        \text{s.t.} \;\; 
        & x_1 + 2x_2 = 0.5, \\
        & 2x_1 - x_2 = 0.2, \\
        & x_1 + x_2 = 0.3
    \end{align*}
\end{subequations}
and its optimization proxy model is learned using \emph{PDL} training.

\paragraph{AC-Optimal Power Flow Problem.}
The OPF determines the least-cost
generator dispatch that meets the load (demand) in a power network.
The OPF is defined in terms of complex numbers, i.e., \emph{powers} of the form $S \!=\! (p \!+\! jq)$, where $p$ and $q$ denote active and reactive powers and $j$ the imaginary unit, \emph{admittances} of the form $Y \!=\! (g \!+\! jb)$, where $g$ and $b$ denote the conductance and susceptance, and \emph{voltages} of the form $V \!=\! (v \angle \theta)$, with magnitude $v$ and phase angle $\theta$. A power network
is viewed as a graph $({\cal N}, {\cal E})$ where the nodes $\cal N$
represent the set of \emph{buses} and the edges $\cal E$ represent
the set of \emph{transmission lines}. The OPF constraints include
physical and engineering constraints, which are captured in the AC-OPF
formulation of Figure \ref{model:ac_opf}.  The model uses $p^g$, and
$p^d$ to denote, respectively, the vectors of active power generation
and load associated with each bus and $p^f$ to describe the vector of
active power flows associated with each transmission line. Similar
notations are used to denote the vectors of reactive power $q$.
Finally, the model uses $v$ and $\theta$ to describe the vectors of
voltage magnitude and angles associated with each bus. The OPF takes
as inputs the loads $(\bm{p}^d\!, \bm{q}^d)$ and the admittance matrix
$\bm{Y}$, with entries $\bm{g}_{ij}$ and $\bm{b}_{ij}$ for each line
$(ij) \!\in\!  {\cal E}$; It returns the active power vector $p^g$ of
the generators, as well the voltage magnitude $v$ at the generator
buses. The problem objective \eqref{c_2a} captures the cost of the
generator dispatch and is typically expressed as a quadratic
function. Constraints \eqref{c_2b} and \eqref{c_2c} restrict the
voltage magnitudes and the phase angle differences within their
bounds.  Constraints \eqref{c_2d} and \eqref{c_2e} enforce the
generator active and reactive output limits.  Constraints \eqref{c_2f}
enforce the line flow limits.  Constraints \eqref{c_2g} and
\eqref{c_2h} capture \emph{Ohm's Law}. Finally, Constraint
\eqref{c_2i} and \eqref{c_2j} capture \emph{Kirchhoff's Current Law}
enforcing flow conservation at each bus. 

\begin{figure}[!t]
\centering
\parbox{.8\linewidth}{\small
    \begin{mdframed}    
    \vspace{-10pt}
    \begin{flalign}
        \minimize: &\hspace{10pt}
        \sum_{i \in {\cal N}}  \text{cost}(p^g_i, \zeta_i) && \label{c_2a} \tag{2a}\\
        \text{s.t.} &\hspace{10pt}
        \bm{v}^{\min}_i \leq v_i \leq \bm{v}^{\max}_i         
                \;\; \forall i \in {\cal N}         \label{c_2b} \tag{2b}\\
        &\hspace{10pt}
        -\bm{\theta}^\Delta_{ij} \leq \theta_i - \theta_j  \leq \bm{\theta}^\Delta_{ij}   
            \;\; \forall (ij) \in {\cal E}       \label{c_2c}\!\!\!\!\! \tag{$2\bar{c}$}\\
        &\hspace{10pt}
        \bm{p}^{g\min}_i \leq p^g_i \leq \bm{p}^{g\max}_i     
            \;\; \forall i \in {\cal N}         \label{c_2d} \tag{$2\bar{d}$}\\
        &\hspace{10pt}
        \bm{q}^{g\min}_i \leq q^g_i \leq \bm{q}^{g\max}_i     
            \;\;\forall i \in {\cal N}         \label{c_2e} \tag{2e}\\
        &\hspace{10pt}
        (p_{ij}^f)^2 + (q_{ij}^f)^2 \leq \bm{S}^{f\max}_{ij}           
            \;\;\forall (ij) \in {\cal E}  \label{c_2f}  \tag{$2\bar{f}$}\\
        &\hspace{10pt}
        p_{ij}^f = \bm{g}_{ij} v_i^2 -  v_i v_j (\bm{b}_{ij} \sin(\theta_i - \theta_j) + \bm{g}_{ij} \cos(\theta_i - \theta_j))   
            \;\;\forall (ij)\in {\cal E} \label{c_2g} \tag{$2\bar{g}$}\\
        &\hspace{10pt}
        q_{ij}^f = - \bm{b}_{ij} v_i^2 -  v_i v_j (\bm{g}_{ij} \sin(\theta_i - \theta_j) - \bm{b}_{ij} \cos(\theta_i - \theta_j))   
            \;\;\forall (ij) \in {\cal E}  \label{c_2h} \tag{2h}\\
        &\hspace{10pt}
            p^g_i - \bm{p}^d_i = \textstyle \sum_{(ij)\in {\cal E}} p_{ij}^f 
            \;\;\forall i\in {\cal N}      \label{c_2i} \tag{$2\bar{i}$}\\
        &\hspace{10pt}
            q^g_i - \bm{q}^d_i = \textstyle  \sum_{(ij)\in {\cal E}} q_{ij}^f    
            \;\;\forall i\in {\cal N}      \label{c_2j} \tag{2j}\\
    \text{Output}:&\hspace{10pt} 
    (p^g, v) \text{ -- The system operational parameters}
    \!\!\!\!\!\!\!\!\!\!\!\!\!\notag
    \end{flalign}
    \vspace{-16pt}
    \end{mdframed}
    }
    \vspace{-6pt}
    \caption{AC Optimal Power Flow (AC-OPF).}
    \label{model:ac_opf}
    \vspace{-12pt}
\end{figure}

\begin{wrapfigure}[5]{r}{175pt}
\vspace{-22pt}
\begin{mdframed}
\vspace{-10pt}
    {\small
    \begin{align}
        \!\!\!\!\!\minimize:& \;\;
        \| p^g - \hat{\bm{p}}^g \|^2 + \| v - \hat{\bm{v}} \|^2 \label{load_flow_obj} 
        \tag{3}\\
        \!\!\!\!\!\text{s.t.:} & \;\; \text{Eqns.}~\ref{c_2b}-\ref{c_2j} \notag\\
        \text{Output}:&\;\;(p^g, v)\notag
    \end{align}
    }
\vspace{-12pt}
\end{mdframed}
\vspace{-6pt}
\caption{AC Load Flow.}
\label{model:load_flow}
\end{wrapfigure}
\setcounter{equation}{3} 
\paragraph{Projection (Load Flow Model)} Being an approximation, a LtO solution $\hat{\bm{p}}^g$ may not satisfy the original constraints. Feasibility can be restored by applying a load flow optimization. A simple load flow
is shown in Figure \ref{model:load_flow}. It is a least square
minimization that finds a feasible solution minimizing the distance to
the approximated one. The use of such a projection allows for
detailed comparison between the various exact and approximate models.
Observe that the load flow itself is a nonlinear nonconvex
problem. However, when started with a good approximation it is
typically much easier to solve than the AC-OPF \cite{fioretto2020predicting}.

\section{Experimental Details}
\label{app:experimental}

\subsection{Portfolio Optimization Dataset}

 The stock return dataset is prepared exactly as prescribed in \cite{sambharya2023end}. The return parameters and asset prices are $\zeta = \alpha(\hat{\zeta}_t + \epsilon_t) $ where $\hat{\zeta}$ is the realized return at time $t$, $\epsilon_t$ is a normal random variable, $\epsilon_t \sim \mathcal{N}(0,\,\sigma_\epsilon I)$, and $\alpha=0.24$ is selected to minimize $\EE \| \hat{\zeta}_t-\zeta \|_2^2$. For each problem instance, the asset prices $\zeta$ are sampled by circularly iterating over the five year interval. In the experiments, see Prob. \ref{eq:opt_portfolio}, $\lambda=2.0$.

The covariance matrix $\bm{\Sigma}$ is constructed from historical price data and set as $\bm{\Sigma}= F\bm{\Sigma}_F F^T + D$, where $F \in \mathbb{R}^{n,l}$ is the factor-loading matrix, $\bm{\Sigma} \in \mathbb{S}_{+}^l$ estimates the factor returns and $D \in \mathbb{S}_{+}^l$, also called the idiosyncratic risk, is a diagonal matrix which takes into account for additional variance for each asset.

\subsection{Hyperparameters}

For all the experiments, the size of the mini-batch $\mathcal{B}$ of the training set is equal to $200$. The optimizer used for the training of the optimization proxy's is Adam, and the learning rate is set to $1e-4$. The same optimizer and learning rate are adopted to train the Two-Stage, EPO (w/o) proxy's predictive model. For each optimization problem, an early stopping criteria based on the evaluation of the test-set precentage regret after restoring feasibility, is adopted to all the LtO(F) the proxies, and the predictive EPO (w/o) proxy. For each optimization problem, an early stopping criteria based on the evaluation of the mean squared error is adopted to all the Two-Stage predictive model.\\  
For each optimization problem, the LtOF proxies are $2$-layers ReLU neural networks with dropout equal to $0.1$ and batch normalization. All the LtO proxies are $(k+1)$-layers ReLU neural networks with dropout equal to $0.1$ and batch normalization, where $k$ denotes the complexity of the feature mapping.
For the LtOF, Two-Stage, EPO (w/o) Proxy algorithm, the feature size of the Convex Quadratic Optimization and Non Convex AC Optimal Power Flow $\lvert z \rvert = 30$, while for the Non Convex Quadratic Optimization $\lvert z \rvert = 50$.
The hidden layer size of the feature generator model is equal to $50$, and the hidden layer size of the LtO(F) proxies, and the 2Stage, EPO and EPO w/ proxy's predictive model is equal to $500$. \\ A grid search method is adopted to tune the  hyperparameters of each LtO(F) models. For each experiments, and for each LtO(F) methods, below is reported the list of the candidate hyperparameters for each $k$, with the chosen ones marked in bold. We refer to \cite{fioretto2020lagrangian}, \cite{park2023self} and \cite{donti2021dc3} for a comprehensive description of the parameters of the LtO methods adopted in the proposed framework. 
In our result, two-stage methods report the \emph{lowest regret} found in each experiment and each $k$ across all hyperparameters adopted, providing a very strong baseline.

\subsubsection{Convex Quadratic Optimization and Non Convex Quadratic Optimization}

\subsection*{LD}
\begin{tabular}{ll}
\toprule
Parameter & Values \\
\midrule
$\lambda$ & $\bm{0.1}$, 0.5, 1.0, 5.0, 10.0, 50.0 \\
$\mu$ & 0.1, $\bm{0.5}$, 1.0, 5.0, 10.0, 50.0 \\
LD step size & 50, 100, $\bm{200}$, 300, 500 \\
LD updating epochs & 1.0, 0.1, 0.01, $\bm{0.001}$, 0.0001 \\
\bottomrule
\end{tabular}

\subsection*{PDL}
\begin{tabular}{ll}
\toprule
Parameter & Values \\
\midrule
$\tau$ & 0.5, 0.6, 0.7, $\bm{0.8}$, 0.9 \\
$\rho$ & 0.1, $\bm{0.5}$, 1, 10 \\
$\rho_{\text{max}}$ & 1000, $\bm{5000}$, 10000 \\
$\alpha$ & 1, 1.5, 2.5, $\bm{5}$, 10 \\
\bottomrule
\end{tabular}

\subsection*{DC3}
\begin{tabular}{ll}
\toprule
Parameter & Values \\
\midrule
$\lambda+\mu$ & 0.1, 1.0, $\bm{10.0}$, 50.0, 100.0 \\
$\frac{\lambda}{\lambda+\mu}$ & 0.1, 0.5, $\bm{0.75}$, 1 \\
$t_{\text{test}}$ & 1, 2, $\bm{5}$, 10, 100 \\
$t_{\text{train}}$ & 1, 2, $\bm{5}$, 50, 100 \\
\bottomrule
\end{tabular}

\subsection*{Non Convex AC-Optimal Power Flow (LD)}
\begin{tabular}{ll}
\toprule
Parameter & Values \\
\midrule
$\lambda$ & $\bm{0.1}$, 0.5, 1.0, 5.0, 10.0, 50.0 \\
$\mu$ & $\bm{0.1}$, 0.5, 1.0, 5.0, 10.0, 50.0 \\
LD step size & 50, 100, 200, $\bm{300}$, 500 \\
LD updating epochs & $\bm{1.0}$, 0.1, 0.01, 0.001, 0.0001 \\
\bottomrule
\end{tabular}










\end{document}

%% file: math_commands.tex
\usepackage{amsmath,amsfonts,bm}

\def\eqref#1{equation~\ref{#1}}

\def\1{\bm{1}}

\DeclareMathAlphabet{\mathsfit}{\encodingdefault}{\sfdefault}{m}{sl}
\SetMathAlphabet{\mathsfit}{bold}{\encodingdefault}{\sfdefault}{bx}{n}

\DeclareMathOperator*{\argmax}{arg\,max}
\DeclareMathOperator*{\argmin}{arg\,min}

%% file: inputs.tex
\usepackage{todonotes}
\usepackage{xcolor}
\usepackage{multirow}
\usepackage{bm} 
\usepackage{paralist}
\usepackage[most]{tcolorbox}
\DeclareMathOperator*{\minimize}{Minimize}

\usepackage{amsmath}
\usepackage{multirow}
\usepackage{multicol}
\usepackage{url}

\newcommand{\rem}[1]{{\color{gray}{#1}}}
\newcommand{\gray}[1]{{\color{gray}{#1}}}

\newcommand{\cC}{\mathcal{C}} \newcommand{\cD}{\mathcal{D}}
 \newcommand{\cF}{\mathcal{F}}
 
 \newcommand{\cL}{\mathcal{L}}

 \newcommand{\cS}{\mathcal{S}}
 
\newcommand{\cZ}{\mathcal{Z}} 
 
 \newcommand{\cX}{\mathcal{X}}

\newcommand{\EE}{\mathbb{E}} \newcommand{\RR}{\mathbb{R}}

\usepackage{esvect}

\makeatletter
\newcommand{\oset}[3][0ex]{\mathrel{\mathop{#3}\limits^{\vbox to#1{\kern-2\ex@\hbox{$\scriptstyle#2$}\vss}}}}
\makeatother

\usepackage{letltxmacro}
\LetLtxMacro\orgvdots\vdots
\LetLtxMacro\orgddots\ddots